\pdfoutput=1
\documentclass[11pt]{article}

\usepackage[]{ACL2023}
\usepackage{times}
\usepackage{latexsym}

\usepackage[T1]{fontenc}
\usepackage[utf8]{inputenc}

\usepackage{microtype}

\usepackage{inconsolata}
\usepackage{graphicx}
\usepackage{multicol}
\usepackage{multirow}
\usepackage{booktabs}
\usepackage{pifont}
\usepackage{xcolor}

\usepackage{amsmath}
\usepackage{xspace}
\usepackage{colortbl}
\usepackage{amsfonts}

\usepackage{enumitem}

\makeatletter
\DeclareRobustCommand\onedot{\futurelet\@let@token\@onedot}
\def\@onedot{\ifx\@let@token.\else.\null\fi\xspace}

\def\eg{\emph{e.g}\onedot} 
\def\ie{\emph{i.e}\onedot} 
 
\def\etc{\emph{etc}\onedot}

\makeatother

\newcommand{\cmark}{\color{blue} \ding{51}}%
\newcommand{\xmark}{\color{red} \ding{55}}%

\newcommand{\method}{\mbox{\textsc{VLM-RLAIF}}\xspace}
\newcommand{\methodsft}{\mbox{\textsc{VLM-SFT}}\xspace}

\definecolor{Gray}{gray}{0.9}

\title{Tuning Large Multimodal Models for Videos using Reinforcement Learning from AI Feedback}

\author{
Daechul Ahn$^{1,3}$\hspace{0.5em}
Yura Choi$^{1,3}$\hspace{0.5em}
Youngjae Yu$^{1}$\hspace{0.5em}
Dongyeop Kang$^{2}$\hspace{0.5em}
Jonghyun Choi$^{3}$\vspace{0.5em}\\
{\hspace{0.5em}$^1$Yonsei University\hspace{0.5em}$^2$University of Minnesota\hspace{0.5em}$^3$Seoul National University} \vspace{0.5em}\\
{\tt\small \{dcahn,yoorachoi,yjy\}@yonsei.ac.kr dongyeop@umn.edu jonghyunchoi@snu.ac.kr}
}

\begin{document}
\maketitle

\begin{abstract}
    Recent advancements in large language models have influenced the development of video large multimodal models (VLMMs). 
    Previous approaches for VLMMs involve Supervised Fine-Tuning (SFT) with instruction-tuned datasets, integrating LLM with visual encoders, and additional learnable parameters.
    Here, aligning video with text, and vice versa, remains a challenge, primarily due to the insufficient quality and quantity of multimodal instruction-tune data compared to that of text-only. 
    This discrepancy often results in alignments that poorly ground the video content. % that are not well-grounded in the video content.
    To address this, we present a novel alignment strategy that employs a multimodal AI system equipped with Reinforcement Learning from AI Feedback (RLAIF), providing self-preference feedback to refine itself and facilitating the alignment of video and text modalities.
    Our approach uniquely integrates detailed video descriptions as context into a multimodal AI system during preference feedback generation to enrich the understanding of video content, a process we call \emph{context-aware reward modeling}.
    Empirical evaluations on various video benchmarks demonstrate that our \method outperforms existing approaches, including the SFT model.
    % Public release of our code, models, and datasets is forthcoming.
    We commit to open-sourcing our code, models, and datasets to foster further research in this area. 
    \url{https://github.com/yonseivnl/vlm-rlaif}
\end{abstract}

\begin{figure}[t]
    \centering
    \includegraphics[width=\linewidth]{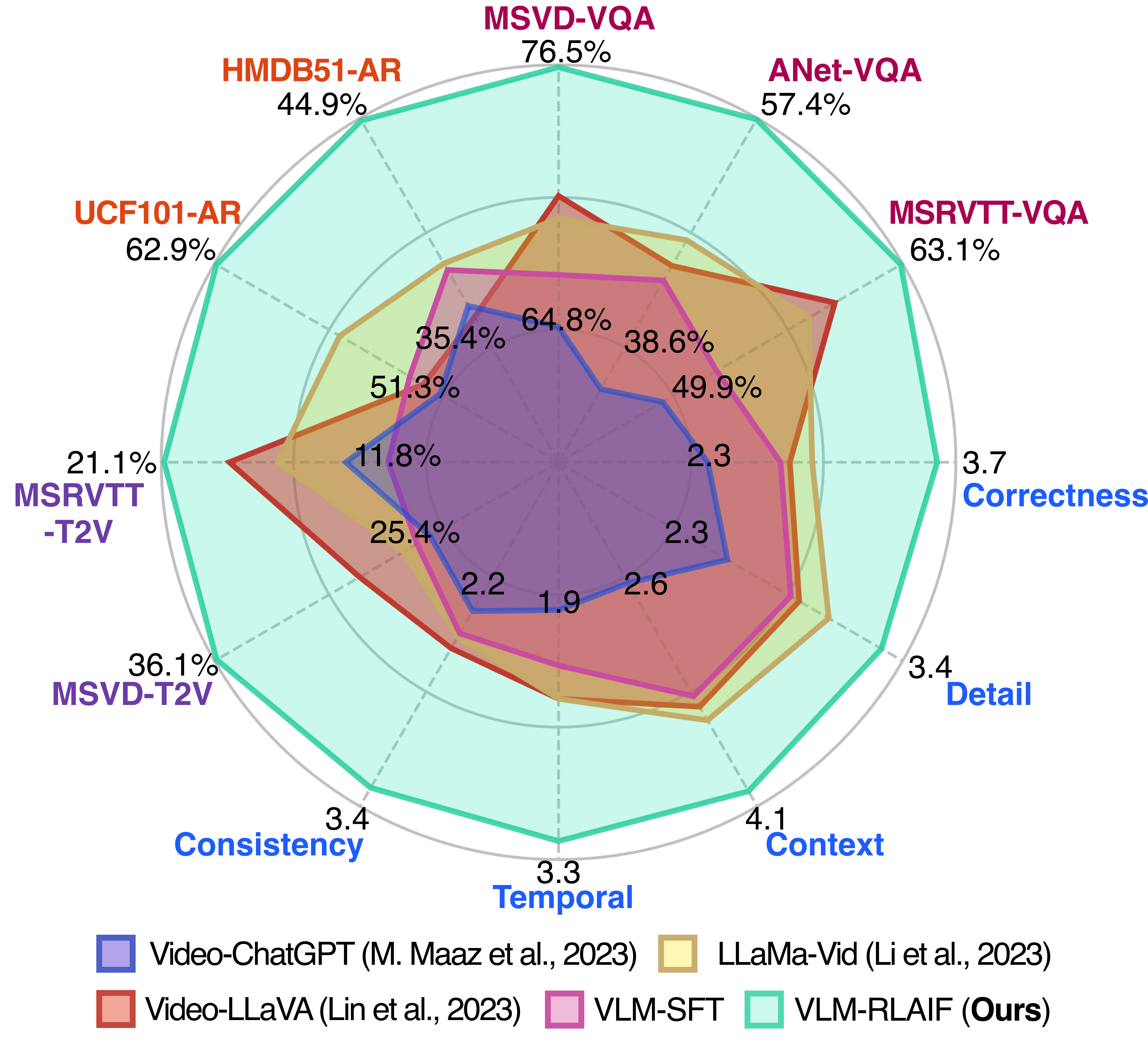}
    \caption{\textbf{Quantitative comparison of VLMMs on various video benchmarks.} The video question answering (VQA) task is marked in \textcolor{purple}{purple}, video-based generative task in \textcolor{blue}{blue}, the text-to-video (T2V) retrieval task in \textcolor{violet}{violet} and the action recognition (AR) task in \textcolor{orange}{orange} color. \method achieves superior performances on a broad range of video benchmarks compared to previous approaches, including \methodsft. Comprehensive comparisons are provided in Tables \ref{tab:vid_bench_main_quan}, \ref{tab:main_quan} and \ref{tab:video_ret_acr_quan}.  
    }
    \label{fig:teaser}
    \vspace{-1em}
\end{figure}

\section{Introduction}
Large language models (LLMs) are advancing many language and multimodal AI tasks, including those involved with video large multimodal models (VLMMs)~\cite{2023videochat,Maaz2023VideoChatGPT,videollava}.
Extending the logical reasoning and advanced cognitive capabilities of LLMs to the visual domain, VLMMs are now remarkably proficient in tasks such as video understanding~\cite{2023videochat}, video question answering~\cite{ko-etal-2023-large} and instruction-following tasks~\cite{Maaz2023VideoChatGPT,luo2023valley}.
These models include publicly available LLMs~\cite{touvron2023llama,vicuna2023,alpaca} with visual encoders and additional learnable parameters~\cite{hu2022lora,liu2023btadapter,blip2}.
To adapt LLMs to the video modality, thus improving their ability to interpret visual content, they all undergo a supervised fine-tuning (SFT) stage using multimodal instruction-tune data~\cite{luo2023valley,Maaz2023VideoChatGPT,2023videochat}.

\begin{figure*}[t]
    \centering
    \includegraphics[width=\linewidth]{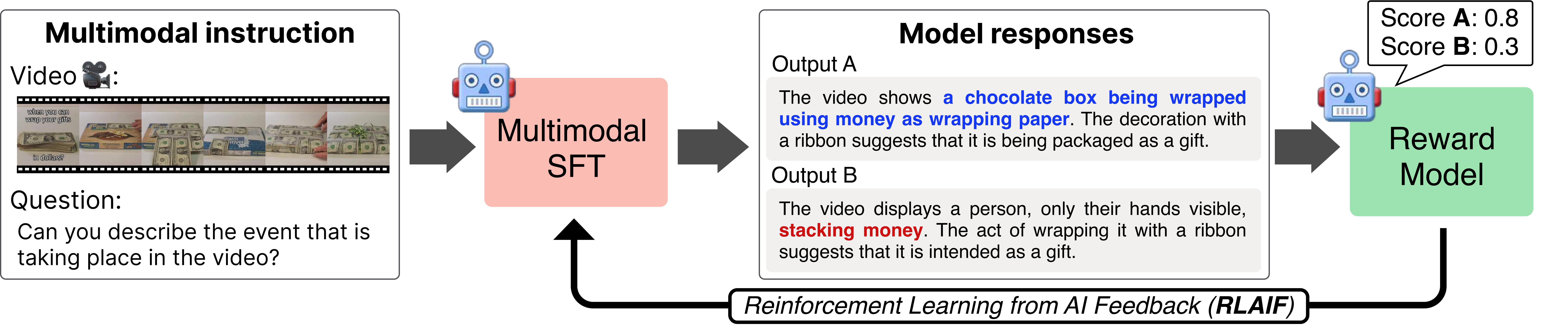}
    \caption{\textbf{Illustration of the proposed VLM-RLAIF.} An LLM tuned with video-text instruction-following data, \ie, multimodal SFT model, often produces responses that are not temporally and visually grounded to the video input, as depicted in \textcolor{red}{red} color. We propose a method that involves using the VLMM to supervise itself by providing self-preference feedback of generated responses using reward model, refining itself and facilitating the alignment of video and text modalities.}
    \label{fig:rlaif_teaser}
\end{figure*}

However, multimodal alignment between video and text faces a significant challenge of deficiency in volume and quality of multimodal instruction-tune data compared to text-only data; text-only data are typically abundant and diverse, while multimodal data are often limited in both quantity and comprehensiveness~\cite{DBLP:journals/corr/abs-2109-01652,liu2023llava}.
This often leads the VLMMs to generate responses that are not properly grounded in the visual content, as illustrated by the SFT model responses in Fig.~\ref{fig:rlaif_teaser}.

To address the issue of VLMMs producing inadequately grounded response, we propose a novel method to align video with text that involves using the VLMM to supervise itself by providing preference feedback of generated responses, as shown in Fig.~\ref{fig:rlaif_teaser}.
Specifically, we propose to use Reinforcement Learning from AI Feedback (RLAIF)~\cite{bai2022constitutional,lee2023rlaif} for multimodal alignment.
Unlike the Reinforcement Learning with Human Feedback (RLHF), which has been successful in aligning text-only or image-text based AI agents with human intentions~\cite{instructgpt,2023llavarlhf,2023rlhf-v}, the RLAIF allows for \emph{scalable oversight} with minimal human intervention.
% Particularly, we enhance AI's feedback by proposing a context-aware reward modeling, focusing on improved clarity in video content (Sec.~\ref{sec:context_pref}).
In particular, we improve AI feedback by leveraging detailed video descriptions as a context during the generation of preference feedback, focusing on improved clarity in video content, a process we refer to as \emph{context-aware reward modeling} (Sec.~\ref{sec:context_pref}).  
In addition, to compensate for the limited multimodal instruction-tune data for training the SFT, we augment it with a human-labeled video question answering and an object-centric multimodal instruction-tune dataset.
Further, to effectively utilize this expanded dataset, we propose a simple curriculum training strategy for enhancing the alignment between video and text modality (Sec.~\ref{sec:sft}). 

We call our proposed framework of training an VLMM with AI feedback as Video Large Multimodal model with RLAIF or \textbf{\method} for short.
Our empirical studies show that our aligned \method exhibits superior performance compared to state-of-the-art VLMMs across a wide array of video benchmarks, as illustrated in Fig.~\ref{fig:teaser}.

We summarize our contributions as follows:
\vspace{-1em}
\begin{itemize}
\setlength\itemsep{-0.2em}
    \item We propose a novel alignment method for video large multimodal models (VLMMs), utilizing Reinforcement Learning from AI feedback (RLAIF) to align video-text modalities effectively.
    \item We enhance AI's feedback by proposing context-aware reward modeling, focusing on improved clarity and comprehension in video.
    \item We enrich the SFT model's training by introducing additional instruction-tune data and applying a simple curriculum strategy. 
    \item We demonstrate the effectiveness of our proposed \method on various video understanding benchmarks by a noticeable margin.
\end{itemize}

\section{Related Work}
\paragraph{Multimodal large model.}
Recently, there have been significant advances for LLMs to go beyond natural language understanding, extending into the realm of multimodal comprehension. 
The goal is to develop LLMs capable of understanding various modalities, \eg, image~\cite{liu2023llava}, video~\cite{2023videochat,videollava}, 3D point-cloud~\cite{guo2023pointbind} and \etc.

To make the LLMs multimodal, most of the work utilize a pretrained encoder, such as CLIP~\cite{radford2021learning}, Q-former~\cite{li2022blip} or ImageBind~\cite{girdhar2023imagebind}, to extract each modality's representations from data. 
These representations are then projected into the token embedding space of the language model.
Then, the models undergo supervised fine-tuning (SFT) with synthetically generated, modality-specific instruction-following datasets.
These approaches, adopted in LLaVA~\cite{liu2023llava}, Video-LLaVA~\cite{videollava} or Point-LLM~\cite{guo2023pointbind}, facilitate the development of proficient conversations grounded in additional modality.

\paragraph{Reinforcement learning from feedback.}
To operate the model safely and in accordance with human intentions, Reinforcement Learning from Human Feedback (RLHF) has been proposed as a viable solution~\cite{instructgpt,2023llavarlhf}.
By collecting preferences from human evaluators, it usually trains the reward model that gives a high reward to the preferred output of the model.
However, a significant challenge in this process is the annotation cost associated with selecting the preference.
To mitigate this issue, Reinforcement Learning from AI Feedback (RLAIF) was proposed~\cite{bai2022constitutional,lee2023rlaif,sun2023salmon}.
RLAIF capitalizes on the inherent ability of Large Language Models (LLMs) to evaluate the generated responses from the SFT model, allowing the LLM itself to assign preferences.

\section{VLM-RLAIF Framework}

% \subsection{Overview}
To overcome the limited scalability of human feedback in RLHF, we use AI's feedback to align multimodality between video and text, reducing the reliance on exhaustive human-annotated preferences~\cite{instructgpt,2023llavarlhf}.
In particular, we improve the feedback process by using detailed video descriptions, thereby achieving better contextual clarity in video content.
The training procedure of \method can be summarized into three stages as follows:

\paragraph{Supervised fine-tuning (SFT).}
We first fine-tune an LLM, \eg, Vicuna, using supervised learning on synthetically generated video-text instruction-tune data~\cite{Maaz2023VideoChatGPT}.
This involves the integration of a vision encoder with two linear layers and additional learnable parameters using LoRA~\cite{hu2022lora}, into the training process.
This fine-tuning allows the model to better follow the instructions~\cite{Maaz2023VideoChatGPT,su2023pandagpt}.
Additionally, we improve the SFT process by expanding the instruction-tune data and introducing simple curriculum learning (Sec.~\ref{sec:sft}).  
We refer to this fine-tuned model as the Video Large Multimodal model with SFT or \textbf{\methodsft} for short.

\paragraph{Reward modeling with AI feedback.}
A key aspect of the RLAIF involves leveraging a pre-trained AI model to generate human-like preferences between different responses generated from the same input~\cite{bai2022constitutional,sun2023salmon,lee2023rlaif}.
To obtain human-like preference, we employ the \methodsft as a judge to assess preferences.
Once preferences are judged, we train a reward model (RM) based on preferences using a cross-entropy loss, following the Bradley-Terry model for estimating score functions from pairwise preferences~\cite{instructgpt,2023llavarlhf}.
We describe the training procedure for collecting preferences and training the reward model in Sec.~\ref{sec:context_pref}.
% We illustrate the training procedure for collecting preferences and training the reward model in the Fig.~\ref{fig:overview}.
The RM give higher score reward to the better response and lower score reward to the less appropriate one in a pair of responses (see examples in Appendix Fig.~\ref{fig:append:reward_viz}), thus guiding the policy model using reinforcement learning (RL).

\paragraph{Reinforcement learning from AI feedback.}
We finally fine-tune a supervised policy model, initialized from the \methodsft, aiming to optimize the scalar reward output of the trained RM by reinforcement learning.
Specifically, we use the Proximal Policy Optimization (PPO) algorithm~\cite{schulman2017ppo}, following~\cite{instructgpt,sun2023salmon,2023llavarlhf}.
% We fine-tune the model to respond optimally to the rewards, thereby enhancing its decision-making abilities in accordance with the learned preferences.

\subsection{Context-Aware Reward Modeling}
\label{sec:context_pref}
For \methodsft to select preference grounded on the video, we argue that a detailed understanding of video content is necessary for more accurate and contextually relevant decisions by the \methodsft.
% Acknowledging this intricacy, it becomes apparent that exclusive dependence on \methodsft for direct video analysis may fall short in achieving thorough comprehension.
% bridging
% Achieving this nuanced understanding of video requires the \methodsft to be aware of the video context.
% We conjecture that because we leverage CLIP image encoder as a video encoder, 
However, the current video encoder presents challenges in accurately encoding the temporal details of videos as they are based on the image encoder~\cite{clip}.

\begin{figure}[t]
    \centering
    \includegraphics[width=\linewidth]{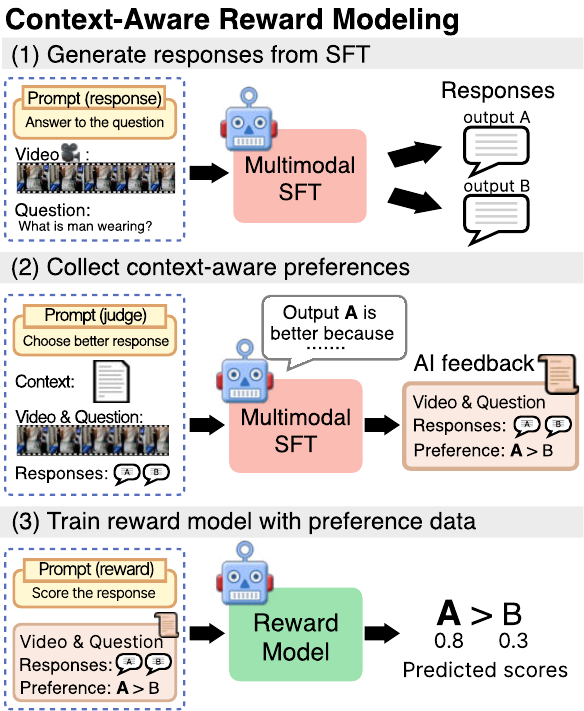}
    \caption{\textbf{The three stages of the proposed context-aware reward modeling.} 
    The work flow of each stage is as follows:
    (1) The SFT model produces two candidate responses from the provided video and question.
    (2) With the video, question and responses at hand, the SFT model utilize context information and guiding prompt to evaluate the responses.
    (3) The RM is trained using the preference pairs generated in the previous step as indicated in orange box.
    Dotted box in each stage denotes a model's input: the first is for generating responses using the SFT model, the second is for the judge model to evaluate and choose the superior response between options A and B, and the third is for training the RM. 
    Each input includes a task-specific prompt, denoted by the yellow box, tailored to guide the model's operation within its respective function (more in Appx. Sec.~\ref{sec:append:input_prompts}).
    }
    \vspace{-1em}
    \label{fig:overview}
\end{figure}

\begin{figure*}[t]
    \includegraphics[width=\linewidth]{figures/Preference_Data_Viz.pdf}
    \caption{\textbf{An example of context-aware preference selection.} We demonstrate our model's ability to generate preference feedback, \ie, preferred choice and explanation marked in orange box, on given instruction input, prompt, two responses and context. \textcolor{red}{Red} color denotes an incorrect response, while \textcolor{blue}{blue} color signifies a correctly grounded response with respect to the visual input. The rationale behind selecting `Output B' as the preferred choice is indicated in \textcolor{teal}{green}.}
    \label{fig:viz_preference}
    \vspace{-1em}
\end{figure*}

\paragraph{Context-aware preference selection.}
We propose to explicitly integrate detailed video descriptions, referred to as \emph{context}, into the preference selection workflow, thereby imparting additional contextual clarity to the VLMM, as illustrated in Figures~\ref{fig:overview}-(2) and~\ref{fig:viz_preference}. 
Specifically, we start by segmenting the video into small clips, each containing up to 20 frames, and then employ the \methodsft to generate a detailed video description for each segment with an input prompt, `Describe this video in detail'.
% we break down the video into smaller clips, then generate descriptive captions for each video segment (refer to Appendix Sec.~\ref{sec:append:gen_context} for more details).
Subsequently, these individual captions are concatenated, which we call a \emph{narrative} of the video.
The narrative is then provided to a judge model, \ie, \methodsft, for better preference selection. 
The context not only improves the \methodsft's ability to obtain a comprehensive view of the video content, but also enables it to identify the most suitable response for the video (see Sec.~\ref{sec:exp:detailed_anal} for empirical results).
Integrating the context with instruction inputs with a specific prompt (rules for generating preferences as illustrated in Appendix Fig.~\ref{fig:append:prompt_judge}), marked in dotted boxes in Fig.~\ref{fig:overview}-(2), allows us to collect context-aware preferences.

\paragraph{Training the reward model.}
We design the reward model (RM) to assign higher scores to responses considered \emph{better} and lower scores to those considered \emph{worse} in quality.
Starting from \methodsft with the final linear layer removed, we train a model to take an input prompt and response, \ie, marked in dotted boxes of Fig.~\ref{fig:overview}-(3), and output a scalar reward value. 
Using the preference dataset produced from the \methodsft, we train the RM with a cross-entropy loss.
Specifically, we use 13B \methodsft to train the reward model, as it gives slightly better performance than 7B \methodsft (see the quantitative comparison discussed in Sec.~\ref{sec:exp:detailed_anal}).
Note that, after training RLAIF using the RM, the 7B \method significantly surpass the 13B \methodsft, thus validating the effectiveness of our proposed framework in aligning video and text modalities (refer to Sec.~\ref{sec:quanti} for more details).

% \subsection{Two-stage Curriculum Supervised Fine-Tuning}
\subsection{Two-stage Curriculum SFT}
\label{sec:sft}
% To obtain video conversational capability, we train the LLM with a open-sourced video-text instruction-tune dataset~\cite{Maaz2023VideoChatGPT}.
During the SFT process, we initially train the LLM with a open-sourced video-text instruction-tune dataset~\cite{Maaz2023VideoChatGPT}.
To improve the VLMM, we not only augment our training with additional video-text instruction-tune datasets but also propose a novel curriculum learning strategy.

\paragraph{Augmenting video instruction-tune dataset.}
To improve the video understanding ability, we first augment the video-text instruction-tune dataset~\cite{Maaz2023VideoChatGPT} with existing human-annotated video question answering datasets~\cite{xiao2021next,li-etal-2020-hero}.
In particular, we focus on obtaining instruction-tune dataset that encompass both visual and temporal intricacies for video comprehension. 
To obtaining such dataset, we use Video Localized Narratives~\cite{Voigtlaender23CVPR}, a dataset that provides detailed descriptions of the appearance and complex actions of all objects in the video.
We transform the narrative dataset into an object-centric instruction-tune dataset format using ChatGPT similar to prior work~\cite{Maaz2023VideoChatGPT}.
More details about the generated instruction-tune dataset are in Appendix Sec.~\ref{sec:append:oc_instruct_tune}.
% More details about the generated instruction-tune dataset can be found in the Appendix Sec.~\ref{sec:append:oc_instruct_tune}.
%We train the SFT model using both the open-sourced and our proposed datasets, resulting in improved performance, as discussed in the Sec.~\ref{sec:exp:detailed_anal}.

\paragraph{Curriculum-based supervised fine-tuning.}
% Recent VLMM research has advanced instruction-following capabilities by curating multimodal instruction-tune data~\cite{Maaz2023VideoChatGPT,luo2023valley}, joint learning across image and video datasets~\cite{videollava,li2023llamavid}, and enhanced bridging networks~\cite{liu2023btadapter,blip2,liu2023llava}.
% Here, we propose to use a structured learning for VLMMs, 
Inspired by the human learning process, we argue the importance of the training sequence, the curriculum learning principles~\cite{curriculum_order} for learning VLMMs. 
We propose a simple two-stage curriculum learning strategy for SFT, aiming to optimize the instruction-following proficiency of VLMMs.

For the curriculum based structured learning, we divide the SFT dataset based on complexity into two segments: \emph{`easy'} and \emph{`hard'}. 
The model is first trained on \emph{`easy'} tasks to learn basic concepts, then progressed to \emph{`hard'} tasks for advanced challenges. 
For the difficulty or easiness of the data, given that longer answers often require deeper comprehension of the context and enhanced proficiency in creating syntactically complex sentences~\cite{xu-etal-2020-curriculum,agrawal-singh-2023-corpus,ranaldi-etal-2023-modeling}, we use \emph{answer length} as our criterion for sample difficulty, \ie, the longer the answer sentence, the more difficult the task is.
% This criterion also allows for a basis for dividing data, offering a concrete measure of difficulty. %that aligns the model's learning trajectory with the incremental nature of human learning.

\begin{table*}[t]
\centering
\resizebox{1.0\linewidth}{!}{
    \begin{tabular}{lcccccc}
    \toprule
    \multirow{2}{*}{\textbf{Methods}}
    & \multirow{2}{*}{\textbf{LLM Size}}
    & \multicolumn{5}{c}{\textbf{Video-based Generative Performance}}  \\
    \cmidrule(lr){3-7}
    &
    & \textbf{Correctness} $\uparrow$ 
    & \textbf{Detail} $\uparrow$  
    & \textbf{Context} $\uparrow$  & \textbf{Temporal} $\uparrow$  & \textbf{Consistency} $\uparrow$  \\
    \midrule
    VideoChat~\cite{2023videochat} & 7B & 2.23 & 2.50 & 2.53 & 1.94 & 2.24 \\
    LLaMA-Adapter~\cite{zhang2023llamaadapter} & 7B & 2.03 & 2.32 & 2.30 & 1.98 & 2.15 \\
    VideoLLaMA~\cite{damonlpsg2023videollama} & 7B & 1.96 & 2.18 & 2.16 & 1.82 & 1.79 \\
    Video-ChatGPT~\cite{Maaz2023VideoChatGPT} & 7B & 2.40 & 2.52 & 2.62 & 1.98 & 2.37 \\
    Valley~\cite{luo2023valley} & 7B & 2.43 & 2.13 & 2.86 & 2.04 & 2.45 \\
    BT-Adapter~\cite{liu2023btadapter} & 7B & 2.68 & 2.69 & 3.27 & 2.34 & 2.46 \\
    VTimeLLM~\cite{huang2023vtimellm} & 7B & 2.78 & 3.10 & 3.40 & 2.49 & 2.47 \\
    Video-LLaVA${^\dagger}$~\cite{videollava} & 7B & 2.84 & 2.86 & 3.44 & 2.46 & 2.57 \\
    VideoChat2~\cite{videochat2} & 7B & 3.02 & 2.88 & 3.51 & 2.66 & 2.81 \\
    LLaMA-VID~\cite{li2023llamavid} & 7B & 2.96 & 3.00 & 3.53 & 2.46 & 2.51 \\
    LLaMA-VID~\cite{li2023llamavid} & 13B & 3.07 & 3.05 & 3.60 & 2.58 & 2.63 \\
    \rowcolor[RGB]{230,230,230}
    GPT-4V~\cite{openai2023gpt4} & - & 3.85 & 3.45 & 3.84 & 3.63 & 2.8 \\
    \midrule
    \methodsft  & 7B & 2.79 & 2.82 & 3.37 & 2.28 & 2.49 \\
    \method  & 7B & \textbf{3.63} & \textbf{3.25} & \textbf{4.00} & \textbf{3.23} & \textbf{3.32} \\
    $\Delta$ (RLAIF - SFT) & - & \textbf{\textit{+0.84}} & \textbf{\textit{+0.43}} & \textbf{\textit{+0.63}} & \textbf{\textit{+0.95}} & \textbf{\textit{+0.83}} \\
    \bottomrule
    \hline
    \end{tabular}
}
\vspace{-0.5em}
\caption{\textbf{Quantitative comparison between different VLMMs on video-based generative performance benchmark.} %We assess various VLMMs on video-based generative benchmark which measures various generative aspects. 
Our approach, \method, shows a performance improvement over previous approaches, with the exception of GPT-4V which requires much more computational resource than ours, and demonstrates noticeable enhancements across five criteria when compared to the \methodsft.
Here, $\Delta$ (RLAIF - SFT) indicates the improvement of RLAIF model over SFT model.
${\dagger}$ denotes reproduced results using the author's implementation.
}
\label{tab:vid_bench_main_quan}
\vspace{-1em}
\end{table*}

\section{Experiments}
\subsection{Experimental Setup}
\paragraph{Model details.}
We initiate training of the \methodsft, building on a pre-trained image-text model~\cite{liu2023llava}, with various video-text instruction-tune datasets. %~\cite{Maaz2023VideoChatGPT,xiao2021next,li-etal-2020-hero,Voigtlaender23CVPR}. 
In particular, we employ a video projection layer consisting of two linear layers with ReLU activation function between them.
Upon establishing the \methodsft, we train the RM using the \methodsft for both its foundation and the generation of preference data. 
Subsequently, we train the RM using reinforcement learning (\method).
The policy model is initialized from the \methodsft, while the value model is initialized from the RM.

\paragraph{Dataset details.}
For the SFT dataset, we utilize the open-source video-text instruction-tune dataset (80k)~\cite{Maaz2023VideoChatGPT,2023videochat} and video question answering datasets (67k)~\cite{xiao2021next,li-etal-2020-hero}.
More importantly, we generate object-centric narrative video-text instruction-tune dataset (180k) for training the \methodsft (Sec.~\ref{sec:sft}).
% In total, we use 327k number of instruction-tune dataset for the SFT. 214 113
For the two-stage curriculum learning, we divide the instruction-tune data for SFT into two groups based on the difficulty; easy (214k) and hard (113k) data.
To train the RM, we first generate responses from existing instruction-tune data~\cite{Maaz2023VideoChatGPT} and generate preferences using them (40k).
Then, we again use the existing instruction-tune dataset (100k)~\cite{Maaz2023VideoChatGPT} for RL with the trained RM.

% \paragraph{Evaluation details.}
% We evaluate our \method on various video benchmarks; zero-shot video questions and video-based generative benchmarks following the prior work~\cite{Maaz2023VideoChatGPT,videollava,li2023llamavid} and text-to-video-retrieval and action recognition as proposed in prior work~\cite{li2023vlmeval}.

\paragraph{Training details.}
For the video input, we uniformly sample 50 frames from each video and extract spatial and temporal features from them using CLIP visual encoder, similar to~\cite{Maaz2023VideoChatGPT}.
%Then, we employ a video projection layer consisting of two linear layers with ReLU activation function between them. % move to model settings
In the two-stage SFT, we set both the LoRA rank and $\alpha$ to 32, respectively, and train the \methodsft for one epoch at each stage.
For RL, we use QLoRA~\cite{dettmers2023qlora}, following~\cite{2023llavarlhf}, setting the rank to 64 and $\alpha$ 16 for computational efficiency and train the policy model for one epoch.
All models are trained using 8$\times$NVIDIA A100 GPUs (80G).
% Additional hyper-parameter settings can be found in the appendix.

\begin{table*}[t]
\centering
\resizebox{1.0\linewidth}{!}{
    \begin{tabular}{lccccccc}
    \toprule
    % & \multicolumn{7}{c}{~~~~~~~~~~~~~~~~~~ Zero-shot Video QA Performance}
    \multirow{2}{*}{\textbf{Methods}}
    & \multirow{2}{*}{\textbf{LLM Size}}
    & \multicolumn{2}{c}{\textbf{MSVD-QA}}
    & \multicolumn{2}{c}{\textbf{MSRVTT-QA}}
    & \multicolumn{2}{c}{\textbf{ActivityNet-QA}}  \\
    % \cmidrule(lr){2-2}   
    \cmidrule(lr){3-4}                  
    \cmidrule(lr){5-6}
    \cmidrule(lr){7-8}
    % \hline
    % & Encoder & T2V 
    % & LLM Size 
    &
    & Acc. & Score 
    & Acc. & Score 
    & Acc. & Score \\
    \midrule
    % \hline
    FrozenBiLM~\cite{yang2022frozenbilm}              
            & 1B & 32.2 & - & 16.8  & - & 24.7 & - \\
    VideoChat~\cite{2023videochat}
            & 7B
            & 56.3 & 2.8 & 45.0 & 2.5 & 26.5 & 2.2 \\
    LLaMA-Adapter~\cite{zhang2023llamaadapter}
            & 7B & 54.9 & 3.1 & 43.8 & 2.7 & 34.2 & 2.7 \\
    VideoLLaMA~\cite{damonlpsg2023videollama}
            & 7B & 51.6 & 2.5 & 29.6 & 1.8 & 12.4 & 1.1 \\
    Video-ChatGPT~\cite{Maaz2023VideoChatGPT}
            & 7B & 64.9 & 3.3 & 49.3 & 2.9 & 35.2 & 2.7 \\
    Valley~\cite{luo2023valley}
            & 7B & 60.5 & 3.3 & 51.1 & 2.9 & 45.1 & 3.2 \\
    BT-Adapter~\cite{liu2023btadapter}
            & 7B & 67.5 & 3.7 & 57.0 & 3.2 & 45.7 & 3.2 \\
    Video-LLaVA~\cite{videollava}
            & 7B & 70.7 & 3.9 & 59.2 & \textbf{3.5} & 45.3 & 3.3 \\
    VideoChat2~\cite{videochat2}
            & 7B & 70.0 & 3.9 & 54.1 & 3.3 & 49.1 & 3.3 \\
    LLaMA-VID~\cite{li2023llamavid}
            & 7B & 69.7 & 3.7 & 57.7 & 3.2 & 47.4 & 3.3 \\
    LLaMA-VID~\cite{li2023llamavid}
            & 13B & 70.0 & 3.7 & 58.9 & 3.2 & 47.5 & 3.3 \\
    \midrule
    \methodsft 
            & 7B & 67.2 & 3.6 & 52.4 & 3.0 & 44.1 & 3.2 \\
    \method 
            & 7B
            & \textbf{76.4} & \textbf{4.0} 
            & \textbf{63.0} & 3.4 
            & \textbf{57.3} & \textbf{3.5} 
            \\
    $\Delta$ (RLAIF - SFT)
            & - 
            & \textbf{\textit{+9.2\%}} & \textbf{\textit{+0.4}}
            & \textbf{\textit{+10.6\%}}   & \textbf{\textit{+0.4}}
            & \textbf{\textit{+13.2\%}} & \textbf{\textit{+0.3}}
            \\
    \bottomrule
    \hline
    \end{tabular}
}
\vspace{-0.5em}
\caption{\textbf{Quantitative comparison between different VLMMs on zero-shot video question answering benchmark.}
\method outperforms previous work across three video-question answering benchmarks.}
\vspace{-0.5em}
\label{tab:main_quan}
% \vspace{-1em}
\end{table*}

\begin{table*}[t]
\centering
\resizebox{1.0\linewidth}{!}{
    \begin{tabular}{lccccccccc}
    \toprule
    \multirow{3}{*}{\textbf{Methods}}
    & \multirow{3}{*}{\textbf{LLM Size}}
    & \multicolumn{4}{c}{\textbf{T2V Retrieval}}
    & \multicolumn{4}{c}{\textbf{Action Recognition}}
    \\
    \cmidrule(lr){3-6}                  
    \cmidrule(lr){7-10}
    &
    & \multicolumn{2}{c}{\textbf{MSVD}}
    & \multicolumn{2}{c}{\textbf{MSRVTT}}
    & \multicolumn{2}{c}{\textbf{UCF101}}
    & \multicolumn{2}{c}{\textbf{HMDB51}}
    \\
    \cmidrule(lr){3-4}                  
    \cmidrule(lr){5-6}
    \cmidrule(lr){7-8}                  
    \cmidrule(lr){9-10}
    &
    & R@1 & R@5 & R@1 & R@5
    & Top-1 & Top-5 & Top-1 & Top-5
    \\
    \midrule
    Video-ChatGPT$^{\dagger}$~\cite{Maaz2023VideoChatGPT}
            & 7B 
            & 26.03 & 51.25 & 14.60 & 33.80 
            & 51.49 & 79.25 & 37.10 & 63.97 \\
    Video-LLaVA$^{\dagger}$~\cite{videollava}
            & 7B 
            & 29.34 & 55.35 & 18.70 & 38.60 
            & 52.33 & 80.86 & 36.64 & 64.03 \\
    LLaMA-VID$^{\dagger}$~\cite{li2023llamavid}
            & 7B 
            & 27.28 & 53.40 & 17.00 & 35.10 
            & 56.58 & 82.79 & 38.85 & 65.27 \\
    \midrule
    \methodsft 
            & 7B 
            & 26.65 & 54.27 & 13.10 & 30.50 
            & 53.03 & 80.34 & 38.58 & 62.37 \\
    \method 
            & 7B 
            & \textbf{36.03} & \textbf{63.40} & \textbf{21.00} & \textbf{40.70} 
            & \textbf{62.83} & \textbf{85.86} & \textbf{44.75} & \textbf{68.37} \\
    $\Delta$ (RLAIF - SFT)
            & - 
            & \textbf{\textit{+9.38}} & \textbf{\textit{+9.13}} & \textbf{\textit{+7.90}} & \textbf{\textit{+10.2}}
            & \textbf{\textit{+9.80}} & \textbf{\textit{+5.52}} & \textbf{\textit{+8.11}} & \textbf{\textit{+6.00}}\\
    \bottomrule
    \hline
    \end{tabular}
}
\vspace{-0.5em}
\caption{\textbf{Quantitative comparison between different VLMMs on zero-shot text-to-video (T2V) retrieval and action recognition.}
Following~\cite{li2023vlmeval}, we evaluate our proposed \method on zero-shot T2V retrieval and action recognition. ${\dagger}$: reproduced by the authors' implementation.}
\label{tab:video_ret_acr_quan}
\vspace{-1em}
\end{table*}

\begin{figure*}[t]
    \centering
    \includegraphics[width=\linewidth]{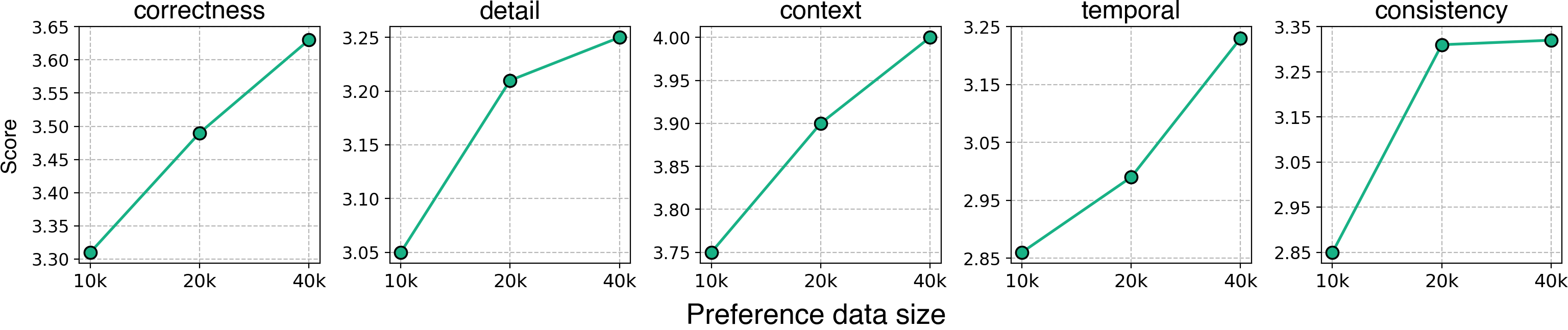}
    \caption{\textbf{Effect of preference data size on video-based generative benchmark.} 
        \method's performance improves across five metrics as the amount of collected preference data increases.
        % Correctness measures relevance to video content, its detail and contextual understanding, temporal understanding, and consistency of prediction in responding to diverse yet related queries.
        The metrics evaluate whether the model generates correct descriptions of the video, includes detailed explanations, remains contextual, demonstrates temporal understanding, and provides consistent responses to variations of the same question.
        % increase with the number of collected preference data.
    }
    \vspace{-1em}
    \label{fig:data_size_gen_bench}
\end{figure*}

\subsection{Quantitative Analysis}
\label{sec:quanti}
We evaluate our proposed \method on various video benchmarks including video-based generative benchmark, zero-shot video question answering~\cite{Maaz2023VideoChatGPT,videollava,li2023llamavid}, text-to-video retrieval, and action recognition~\cite{li2023vlmeval}.

\paragraph{Video-based generative performance.}
We evaluate VLMMs on the video-based generative performance benchmark~\cite{Maaz2023VideoChatGPT} that measures five criteria of generated text. %: correctness, detail, context, temporal understanding, and consistency of generated text. 
In specific, these assess the relevance of the model's output to the video content, its capacity to capture essential details and contextual information, its understanding of temporal sequences, and the consistency in responding to varied yet related queries.
As shown in Tab.~\ref{tab:vid_bench_main_quan}, the \method performs \emph{on par with} GPT-4V~\cite{openai2023gpt4}, which requires much more computational resources than ours (\ie, not a fair comparison), and outperforms previous approaches and the \methodsft.
% We believe that aligning video-text modalities through RLAIF improves the overall generative capabilities.

\paragraph{Zero-shot video question answering.}
%To assess the general capabilities of VLLMs, we conduct a quantitative evaluation of their video question answering (VideoQA) abilities using three datasets: MSVD-QA, MSRVTT-QA~\cite{xu2017videoqa}, and ActivityNet-QA~\cite{yu2019activityqa}.
To evaluate the reasoning ability of VLLMs, we conduct a quantitative evaluation of video question answering (VideoQA) abilities on three datasets~\cite{xu2017videoqa,yu2019activityqa}, following~\cite{Maaz2023VideoChatGPT}.
% We adopted the evaluation pipeline from prior work~\cite{Maaz2023VideoChatGPT} to gauge the zero-shot VideoQA performance.

\begin{figure}[!t]
    \includegraphics[width=\linewidth]{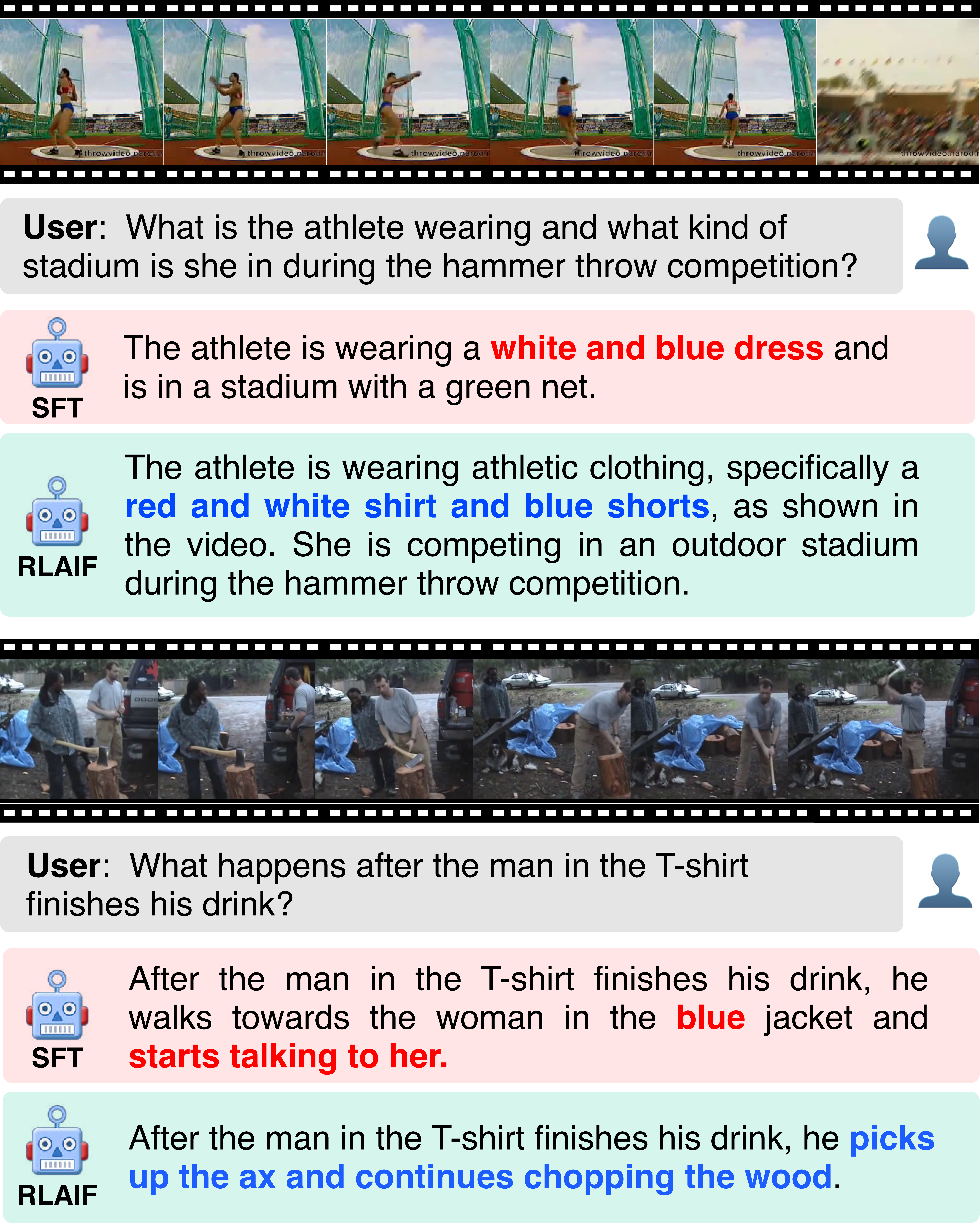}
    \caption{\textbf{Samples of \method and \methodsft in video understanding.} Two samples demonstrate better capability of the \method. \textcolor{red}{Red} color indicates visually un-grounded expressions considering video content and \textcolor{blue}{blue} color indicates well grounded expressions to the video.}
    \label{fig:quali_sft_rlaif}
    \vspace{-1em}
\end{figure}

The results, as shown in Table \ref{tab:main_quan}, indicate that the \method significantly outperforms previous approaches, including \methodsft. 
Notably, \method exceeds \methodsft by 9.2\%, 10.6\%, and 13.2\% in accuracy and by 0.4, 0.4, and 0.3 in score across all datasets.
We believe that the better visually-aligned response generated from the \method improves the performance (see quantitative analysis in Figures~\ref{fig:quali_sft_rlaif} and \ref{fig:append:more_sft_rlaif}).

\paragraph{Zero-shot text-to-video retrieval.}
For this task, we follow the procedure proposed in~\cite{li2023vlmeval}, which compares CLIP score between generated description and ground-truth caption.
Table~\ref{tab:video_ret_acr_quan} illustrates the summarized performance comparison to various VLMMs.
In the two datasets, \ie, MSVD and MSRVTT, the proposed \method clearly outperforms other methods including our \methodsft by the help of better alignment by the proposed components.

\paragraph{Zero-shot action recognition.}
Following the VLMMs evaluation procedure proposed in~\cite{li2023vlmeval}, we conduct zero-shot action recognition task using two benchmark datasets, \eg, UCF101 and HMDB51.
We summarize results of various VLMMs in Tab.~\ref{tab:video_ret_acr_quan}.
In two datasets, the proposed \method again clearly outperforms other methods including our \methodsft.

\begin{table}[t]
\centering
\resizebox{1.0\linewidth}{!}{
    \begin{tabular}{ccccccccc}
    \toprule
    \multicolumn{3}{c}{\textbf{SFT datasets}}
    & \multicolumn{1}{c}{\multirow{2}{*}{\shortstack{\textbf{Curr.}\\\textbf{learning}}}}
    & \multicolumn{5}{c}{\textbf{Video-based Generative Performance}} 
    \\
    \cmidrule(lr){1-3}
    \cmidrule(lr){5-9}
    [A] & [B] & [C] &
    & \textbf{Corr.} $\uparrow$ 
    & \textbf{Det.} $\uparrow$ 
    & \textbf{Cont.} $\uparrow$ 
    & \textbf{Temp.} $\uparrow$ 
    & \textbf{Cons.} $\uparrow$ 
    \\
    \midrule
        \cmark & \xmark & \xmark & \xmark
        & 2.32 & 2.53 & 3.03 & 2.16 & 2.23 \\
        \cmark & \cmark & \cmark & \xmark 
        & 2.43 & 2.56 & 3.09 & 2.19 & 2.19 \\
        \cmark & \cmark & \cmark & \cmark 
        & \textbf{2.79} & \textbf{2.82} & \textbf{3.37} & \textbf{2.28} & \textbf{2.49} \\
    \bottomrule
    \hline
    \end{tabular}
}
\vspace{-0.5em}
\caption{\textbf{In-depth analysis for the \methodsft training procedure.} `[A]' indicate the multimodal instruction-tune dataset proposed in~\cite{Maaz2023VideoChatGPT,2023videochat}. `[B]' represents the use of a human-labeled video question answering dataset~\cite{xiao2021next,li-etal-2020-hero}, while `[C]' refers to the use of an object-centric video narrative instruction-tune dataset (Appendix Sec.~\ref{sec:append:oc_instruct_tune}). `Curr. learning' indicates the curriculum learning (Sec.~\ref{sec:sft}).}
\label{tab:sft_dataset}
\vspace{-1em}
\end{table}

\begin{table}[t]
\centering
\resizebox{1.0\linewidth}{!}{
    \begin{tabular}{cccccccc}
    \toprule
    \multirow{2}{*}{\textbf{RLAIF}}
    & \multirow{2}{*}{\textbf{Context}}
    & \multirow{2}{*}{$\#$\textbf{Clips}}
    & \multicolumn{5}{c}{\textbf{Video-based Generative Performance}} \\
    % & \cmidrule(lr){4-8}
    \cmidrule(lr){4-8}
    & \textbf{Info.} & 
    & \textbf{Corr.} $\uparrow$ 
    & \textbf{Det.} $\uparrow$ 
    & \textbf{Cont.} $\uparrow$ 
    & \textbf{Temp.} $\uparrow$ 
    & \textbf{Cons.} $\uparrow$ 
    \\
    \midrule
    \xmark
            & \xmark & -
            & 2.79 & 2.82 & 3.37 & 2.28 & 2.49 \\ 
    \cmark
            & \xmark & -
            & 3.26 & 3.11 & 3.74 & 2.78 & 3.14 \\ 
    \cmark
            & \cmark & 1
            & 3.44 & 3.20 & 3.89 & 2.97 & \textbf{3.36} \\ 
    \cmark
            & \cmark & 3
            & \textbf{3.63} &\textbf{3.25} &\textbf{4.00} &\textbf{3.23} & 3.32 \\ 
    \bottomrule
    \hline
    \end{tabular}
}
\vspace{-0.5em}
\caption{\textbf{Effect of context information on video-based generative performance benchmark.} We investigate the efficacy of using context information for reward modeling (Sec.~\ref{sec:context_pref}).
`Context Info.' indicates the use of context in preference selection.
`$\#$ Clips' denotes the number of segments into which we divide the video to generate the context information. }
\label{tab:context_info}
\vspace{-1em}
\end{table}

\begin{table}[th]
\centering
\resizebox{1\linewidth}{!}{
    \begin{tabular}{lcccccc}
    \toprule
    \multirow{2}{*}{\textbf{Methods}}
    & \multirow{2}{*}{\textbf{LLM Size}}
    & \multicolumn{5}{c}{\textbf{Video-based Generative Perf.}}
    \\
    \cmidrule(lr){3-7} 
    & & \textbf{Corr.} $\uparrow$  & \textbf{Det.} $\uparrow$  & \textbf{Cont.} $\uparrow$  & \textbf{Temp.} $\uparrow$  & \textbf{Cons.} $\uparrow$  \\
    \midrule
    \methodsft 
            & 7B
            & 2.32 & 2.53 & 3.03 & 2.16 & 2.23 \\
    \methodsft
            & 13B 
            & \textbf{2.64} & \textbf{2.73} & \textbf{3.28} & \textbf{2.38} & \textbf{2.44} \\
    \bottomrule
    \end{tabular}
}
\caption{
\textbf{Quantitative comparison between different sizes of the \methodsft.} We assess the performance of \methodsft with varying LMM sizes, specifically 7B and 13B, on video-based generative benchmarks. We conduct this evaluation without the integration of augmented instruction-tune data and the implementation of two-stage curriculum learning.}
\label{tab:append:sft_7b_13b}
\end{table}

\begin{table*}[th]
\centering
\resizebox{1.0\linewidth}{!}{
    \begin{tabular}{lcccccccccccccc}
    \toprule
    \multirow{3}{*}{\textbf{Methods}}
    & \multicolumn{2}{c}{\textbf{LLM Size}}
    & \multicolumn{6}{c}{\textbf{Video Question Answering}}
    & \multicolumn{5}{c}{\textbf{Video-based Generative Perf.}}
    \\
    % \cmidrule{lr}{2-3}
    \cmidrule(lr){2-3}
    \cmidrule(lr){4-9} 
    \cmidrule(lr){10-14}
    & \multirow{2}{*}{Policy}
    & \multirow{2}{*}{Reward}
    & \multicolumn{2}{c}{\textbf{MSVD}}
    & \multicolumn{2}{c}{\textbf{MSRVTT}}
    & \multicolumn{2}{c}{\textbf{ActivityNet}}
    & \multirow{2}{*}{\textbf{Corr.} $\uparrow$ } & \multirow{2}{*}{\textbf{Det.} $\uparrow$ } & \multirow{2}{*}{\textbf{Cont.} $\uparrow$ } & \multirow{2}{*}{\textbf{Temp.} $\uparrow$ } & \multirow{2}{*}{\textbf{Cons.} $\uparrow$ } 
    \\
    \cmidrule(lr){4-5}                  
    \cmidrule(lr){6-7}
    \cmidrule(lr){8-9}        
    & Model & Model & Acc. & Score & Acc. & Score & Acc. & Score & & 
    \\
    \midrule
    \rowcolor[RGB]{230,230,230}
    LLaMA-VID (7B)
            & - & -
            & 69.7 & 3.7 & 57.7 & 3.2 & 47.4 & 3.3 
            & 2.96 & 3.00 & 3.53 & 2.46 & 2.51 \\
    \midrule
    \methodsft 
            & 7B & -
            & 67.2 & 3.6 & 52.4 & 3.0 & 44.1 & 3.2 
            & 2.79 & 2.82 & 3.37 & 2.28 & 2.49 \\
    \method 
            & 7B & 7B 
            & 75.1 & 3.9 & 61.0 & 3.3 & 56.1 & 3.4
            & 3.47 & 3.14 &	3.87 & 3.05 & 3.30 \\
    \method
            & 7B & 13B 
            & \textbf{76.4} & \textbf{4.0} & \textbf{63.0} & \textbf{3.4} & \textbf{57.3} & \textbf{3.5}
            & \textbf{3.63} & \textbf{3.25} & \textbf{4.00} & \textbf{3.23} & \textbf{3.32} \\
    \bottomrule
    \hline
    \end{tabular}
}
% \caption{
% \textbf{Quantitative comparison between different policy and reward model sizes for \method on zero-shot video question answering and video-based generative benchmark.}}
\caption{
\textbf{Quantitative comparison between different size of policy model and reward model for the \method.} We evaluate the \method with different model size for policy model and reward model on zero-shot video question answering and video-based generative benchmark.}
\label{tab:append:rm_7b_13b_vqa}
\end{table*}

\subsection{Detailed Analysis}
\label{sec:exp:detailed_anal}
For a detailed analysis, we use the video-based generative benchmark~\cite{Maaz2023VideoChatGPT} specifically, as it is well suited to evaluate the wide-ranging capabilities of VLMM, \ie, focusing on response relevance, detail and context capture, temporal understanding, and consistency across queries.

\paragraph{In-depth analysis of SFT training.}
We first empirically support the effectiveness of augmenting the SFT dataset with additional instruction-following dataset (Sec.~\ref{sec:sft}).
The first and second rows of Tab.~\ref{tab:sft_dataset} illustrate the benefits of incorporating this additional dataset in improving performance. 
On top of that, the application of curriculum learning significantly improves performance, implying the efficacy of curriculum learning for the SFT process (the third row of Tab.~\ref{tab:sft_dataset}).

\paragraph{Effect of preference data size.}
Our method's strength lies in generating synthetic preference feedback in large quantities. 
To study the benefit of large-sized data, we sweep the size of preference data from 10k -- the same quantity utilized in the multimodal image-text RLHF framework~\cite{2023llavarlhf} -- to 40k. 
As expected, we observe monotonic increases in performance by the increase in data size, as shown in Figure~\ref{fig:data_size_gen_bench}.

\begin{table*}[!th]
\centering
\resizebox{1.0\linewidth}{!}{
    \begin{tabular}{lcccccccccccc}
    \toprule
    \multirow{3}{*}{\textbf{Methods}}
    & \multicolumn{2}{c}{\textbf{LLM Size}}
    & \multicolumn{4}{c}{\textbf{T2V Retieval}}
    & \multicolumn{4}{c}{\textbf{Action Recognition}}
    \\
    % \cmidrule{lr}{2-3}
    \cmidrule(lr){2-3}
    \cmidrule(lr){4-7}                  
    \cmidrule(lr){8-11}
    & \multirow{2}{*}{Policy}
    & \multirow{2}{*}{Reward}
    & \multicolumn{2}{c}{\textbf{MSVD}}
    & \multicolumn{2}{c}{\textbf{MSRVTT}}
    & \multicolumn{2}{c}{\textbf{UCF101}}
    & \multicolumn{2}{c}{\textbf{HMDB51}}
    \\
    \cmidrule(lr){4-5}                  
    \cmidrule(lr){6-7}
    \cmidrule(lr){8-9}                  
    \cmidrule(lr){10-11}
    & Model & Model & R@1 & R@5 & R@1 & R@5
    & Top-1 & Top-5 & Top-1 & Top-5
    \\
    \midrule
    \rowcolor[RGB]{230,230,230}
    LLaMA-VID (7B)
            & - & -
            & 27.28 & 53.40 & 17.00 & 35.10 
            & 56.58 & 82.79 & 38.85 & 65.27 \\
    \midrule
    \methodsft 
            & 7B & -
            & 26.65 & 54.27 & 13.10 & 30.50 
            & 53.03 & 80.34 & 38.58 & 62.37 \\
    \method 
            & 7B & 7B 
            & 33.73 & 61.95 & 20.80 & \textbf{42.90}
            & 61.09 & 85.15 & 43.86 & 65.88 \\
            
    \method
            & 7B & 13B 
            & \textbf{36.03} & \textbf{63.40} & \textbf{21.00} & 40.70 
            & \textbf{62.83} & \textbf{85.86} & \textbf{44.75} & \textbf{68.37} \\
    \bottomrule
    \hline
    \end{tabular}
}
% \caption{
% \textbf{Quantitative comparison of different policy and reward model sizes for \method on zero-shot text-to-video retrieval and action recognition tasks.}}
\caption{
\textbf{Quantitative comparison between different size of policy model and reward model for the \method.} We evaluate the \method with different model size for policy model and reward model on zero-shot text-to-video retrieval and action recognition tasks.}
\label{tab:append:rm_7b_13b_ret}
\end{table*}

% \begin{figure}[!t]

\paragraph{Context-aware preference labeling.}
Table~\ref{tab:context_info} presents a study on the impact of context information. %, as discussed in Sec.~\ref{sec:context_pref}.
This reveals that the use of context information during preference selection improves performance, as demonstrated in the second and third rows in Tab.~\ref{tab:context_info}.
Moreover, segmenting the video into multiple clips, specifically three, and using their aggregated captions as context further improves performance (see the third and fourth rows). 
We conjecture that segmenting video into clips with detailed captions enables the model to capture detailed nuances, thereby enriching the contextual clarity for reward modeling. % and boosting the performance.

\paragraph{Comparison between different SFT model sizes.}
Table~\ref{tab:append:sft_7b_13b} shows a clear performance margin between \methodsft with different LLM sizes, specifically 7B and 13B, in video-based generative tasks.
Larger LLM sizes lead to improved performance, suggesting that increased model capacity enhances the ability to capture and generate complex video content.
Thus, we adopt the 13B model for the reward model (Sec.~\ref{sec:context_pref}), with supporting evidence in Tables \ref{tab:append:rm_7b_13b_vqa} and \ref{tab:append:rm_7b_13b_ret}.
% Table~\ref{tab:append:sft_7b_13b} shows clear performance margin between various \methodsft using varying LLM sizes, specifically 7B and 13B in video-based generative tasks, \ie, larger LLM sizes lead to improved performance.
% This suggests that the increased model capacity inherent in larger LLMs enhances the model's ability to capture and generate more complex and nuanced video content representations.
% Thus, we adopt the 13B model for the reward model (Sec.~\ref{sec:context_pref}), with empirical evidence provided in Tables \ref{tab:append:rm_7b_13b_vqa} and \ref{tab:append:rm_7b_13b_ret}.

\paragraph{Various LLM size for reward model}
Tables~\ref{tab:append:rm_7b_13b_vqa} and \ref{tab:append:rm_7b_13b_ret} show the performance of \method with different RM sizes, initialized from \method-7B and \method-13B.
In both cases, the policy model is initialized with \methodsft-7B.
Our RLAIF method outperforms \methodsft significantly across all evaluation methods. 
Specifically, RLAIF with 7B RM achieves a 5-12\% improvement in zero-shot video question answering. 
Scaling up the RM from 7B to 13B further improves performance, except for the text-to-video retrieval task R@5 metric on MSR-VTT.

\subsection{Qualitative Analysis}
% \paragraph{\methodsft \textit{vs}. \method.}
We now qualitatively compares the performance of \methodsft and \method, highlighting their multimodal understanding capabilities in Figure~\ref{fig:quali_sft_rlaif}. 
\method consistently yields more accurate answers than \methodsft, as shown in the detailed recognition of the attire of an athlete in the first example, marked in \textcolor{red}{red} and \textcolor{blue}{blue}. 
The second example further affirms \method's benefit in generating better grounded responses to the visual input, where \methodsft falls short.
More examples are in the Appendix Fig.~\ref{fig:append:more_sft_rlaif} for the space sake.

\section{Conclusion}
We propose a novel alignment strategy for VLMMs, termed \method, that uses reinforcement learning from AI feedback. 
To improve multimodal alignment, we propose a context-aware reward modeling, enabling AI to generate feedback for self-improvement. 
In addition, we expand the instruction-tune dataset for SFT and adopt a curriculum-based training approach, that is particularly effective in the gradual learning of complex video-text relationships.
In our empirical validations, the \method significantly outperforms previous models in multiple multimodal video-text understanding benchmarks, which implies good generalization performance across tasks.

\section*{Limitations}
% Relibaility of preference feedback from the SFT model
% We propose a novel alignment approach for video-text modalities through reinforcement learning from AI feedback (RLAIF) without relying on human preference feedback.
Given that our approach utilizes feedback synthesized by the AI model, the effectiveness of our proposed \method largely depends on the quality of the AI model's generated responses.
In light of recent studies exploring the use of artificially generated data~\cite{koo2023benchmarking,das2024surface}, we believe that there needs further research to enhance the quality of synthetically generated data, thereby establishing a more reliable RLAIF system.  
% Additional tasks

In addition, although we have evaluate our model across a range of benchmarks for VLMMs, \eg, videoQA, video-based generative tasks, retrieval, and recognition tasks, there are other tasks such as temporal reasoning~\cite{liang2022var} that are necessary for VLMMs to be effectively applied in real-world scenarios.
Application of our method to these tasks would be a great future research avenue.

% Considering recent works~\cite{koo2023benchmarking,das2024surface} that investigate the potential of artificially generated data, we believe that more investigation for improving the reliability in synthetically generated data is necessary, thereby implementing reliable RLAIF system.
% Thus, Our \method heavily relies on the quality of the synthetically generated feedback from the SFT model.
% We believe that the reliability of the SFT model 
% Hence, a further study is necessary to investigate the effect of preference feedbaack's in order to ... 
% Even we improve the VLMM's performance with the AI feedback process, our \method rely on the quality of the preference feedback from the SFT model.
% Hence, if there are 
% Need to evaluate the model on various video benchmarks
% In addition, even though we evaluate our model on various proposed VLMM's benchamrks~\cite{}, \eg, videoQA, video-based generative, retrieval and recognition tasks, we believe more evaluation may not enough to evaluate 

\section*{Acknowledgment}
This work was partly supported by the NRF grant (No.2022R1A2C4002300, 20\%) and IITP grants (No.RS-2022-II220077 (10\%), No.RS-2022-II220113 (25\%), No.RS-2022-II220959 (10\%), No.RS-2022-II220871 (10\%), No.RS-2022-II220113 (10\%, Yonsei AI), No.RS-2021-II211343 (5\%, SNU AI), No.RS-2020-II201361 (5\%, AI Innov. Hub), No.RS-2022-11220951 (5\%)) funded by the Korea government (MSIT), NCSOFT and Artificial intelligence industrial convergence cluster development project funded by MSIT and Gwangju Metropolitan City.
% We thank anonymous reviewers for their comments.
% This work is supported in part by Artificial intelligence industrial convergence cluster development project funded by the Ministry of Science and ICT (MSIT, Korea) and Gwangju Metropolitan City.

\bibliography{anthology,custom}
\bibliographystyle{acl_natbib}

\clearpage
\appendix

\section*{Appendix}

\section{Details About Object-Centric Instruction-Tune Data}
\label{sec:append:oc_instruct_tune}
To enhance the alignment of Large Language Models (LLMs) with video content, we prioritize acquiring more video-text instruction-tune dataset that captures both the visual and temporal complexities of videos. 
For this, we leverage the Video Localized Narratives dataset~\cite{Voigtlaender23CVPR}, which contains a comprehensive description of each object's appearance and action, along with their tracked coordinates within videos.
To utilize the best of the rich semantics contained in narrative dataset, we feed all ground-truth captions to the model, \ie, ChatGPT, and prompt it to generate question and answer pairs, as shown in the template in Fig.~\ref{fig:append:if_data_gen_template}-(a).
We aim to transform all object's descriptions into instruction-tune data format, \ie, question and answer pair, which demands an understanding of the visual specifics of each object and its surroundings, ensuring the questions and answers are anchored in the video. 
An example of the generated instruction-tune data is depicted in Fig.\ref{fig:append:if_data_gen_template}-(b), showcasing the approach's effectiveness in creating contextually rich instructional content.

\section{Input Prompts for Reward Modeling}
\label{sec:append:input_prompts}

\begin{figure*}[t]
    \includegraphics[width=\linewidth]{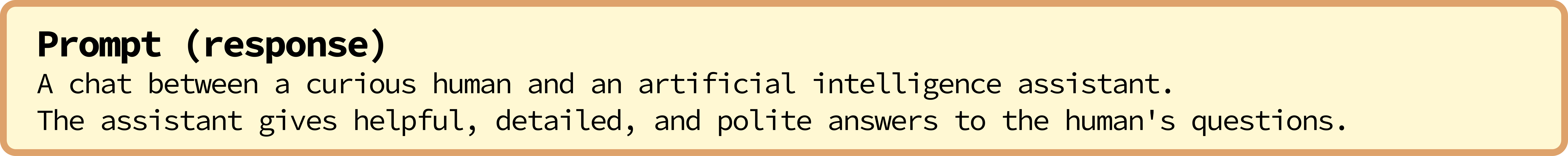}
    \caption{\textbf{Designed input prompt fed to an \methodsft to generate responses.} }
    \label{fig:append:prompt_responses}
\end{figure*}

We leverage three types of input prompts for context-aware reward modeling, as discussed in Sec.~\ref{sec:context_pref}.
Figure~\ref{fig:append:prompt_responses} presents a designed input prompt given to our method, resulting in two responses. 
Furthermore, Figure~\ref{fig:append:prompt_judge} illustrates the detailed input prompt used to select a preference between two responses. Additionally, Figure~\ref{fig:append:prompt_reward} displays the input prompt employed for generating a reward score.

% \section{Details About Generating Contextual Information}
% \label{sec:append:gen_context}
% For effective utilization of \methodsft in preference selection, a nuanced understanding of video is essential.
% To achieve this, we equip the \methodsft with detailed video descriptions, referred to as `context', during preference selection process, to enhance contextual clarity in the video.
% Specifically, we start by segmenting the video into small clips, each containing up to 20 frames, and then employ the \methodsft to generate a detailed video description for each segment with an input prompt, `Describe this video in detail'.
% We then aggregate the descriptions of each segment by concatenating them, thereby generating \emph{context} of the video.

% \section{Various LLM Size for Reward Model}
% In Tables~\ref{tab:append:rm_7b_13b_vqa} and \ref{tab:append:rm_7b_13b_ret}, we assess the performance of \method using varying reward model(RM) sizes, each initialized from \method-7B and \method-13B.
% In both settings, we use 7B \methodsft to initialize the policy model.
% For both sizes of RM, our RLAIF method showed a performance gain with a large margin from \methodsft across all evaluation methods.
% Specifically, RLAIF with 7B RM acquired 5-12\% improvement on zero-shot video question answering.
% Scaling up the RM from 7B to 13B result in more improved performance, except for text-to-video retrieval task R@5 metric on MSR-VTT.

\begin{figure*}[t]
    \centering
    \includegraphics[width=0.94\linewidth]{figures/SFT_NewData-reduced.pdf}
    \caption{\textbf{Designed input prompt for object-centric instruction-tune dataset generation from video narrative dataset and the data sample.} 
    In (a), the prompt we used to generate instruction-following dataset is displayed in a white box.
    It includes the system prompt, a task definition, guidelines, principles, and an example set of input and desired output.
    Using this prompt with ChatGPT-3.5-turbo, we create an additional instruction following datasets.
    The resulting example is visualized in (b).
    The question covers details of the video scenes, such as the number and appearance of parrots and the action of each parrot.
    }
    \label{fig:append:if_data_gen_template}
\end{figure*}

\clearpage

\begin{figure*}[t]
    \includegraphics[width=\linewidth]{figures/prompt_judge.pdf}
    \caption{\textbf{Designed input prompt fed to an \methodsft to generate AI's preference feedback.} }
    \label{fig:append:prompt_judge}
\end{figure*}

\begin{figure*}[t]
    \includegraphics[width=\linewidth]{figures/prompt_reward.pdf}
    \caption{\textbf{Designed input prompt fed to a reward model to generate scalar reward score.} }
    \label{fig:append:prompt_reward}
\end{figure*}

\clearpage

\begin{figure*}[t]
    % \centering
    \includegraphics[width=\linewidth]{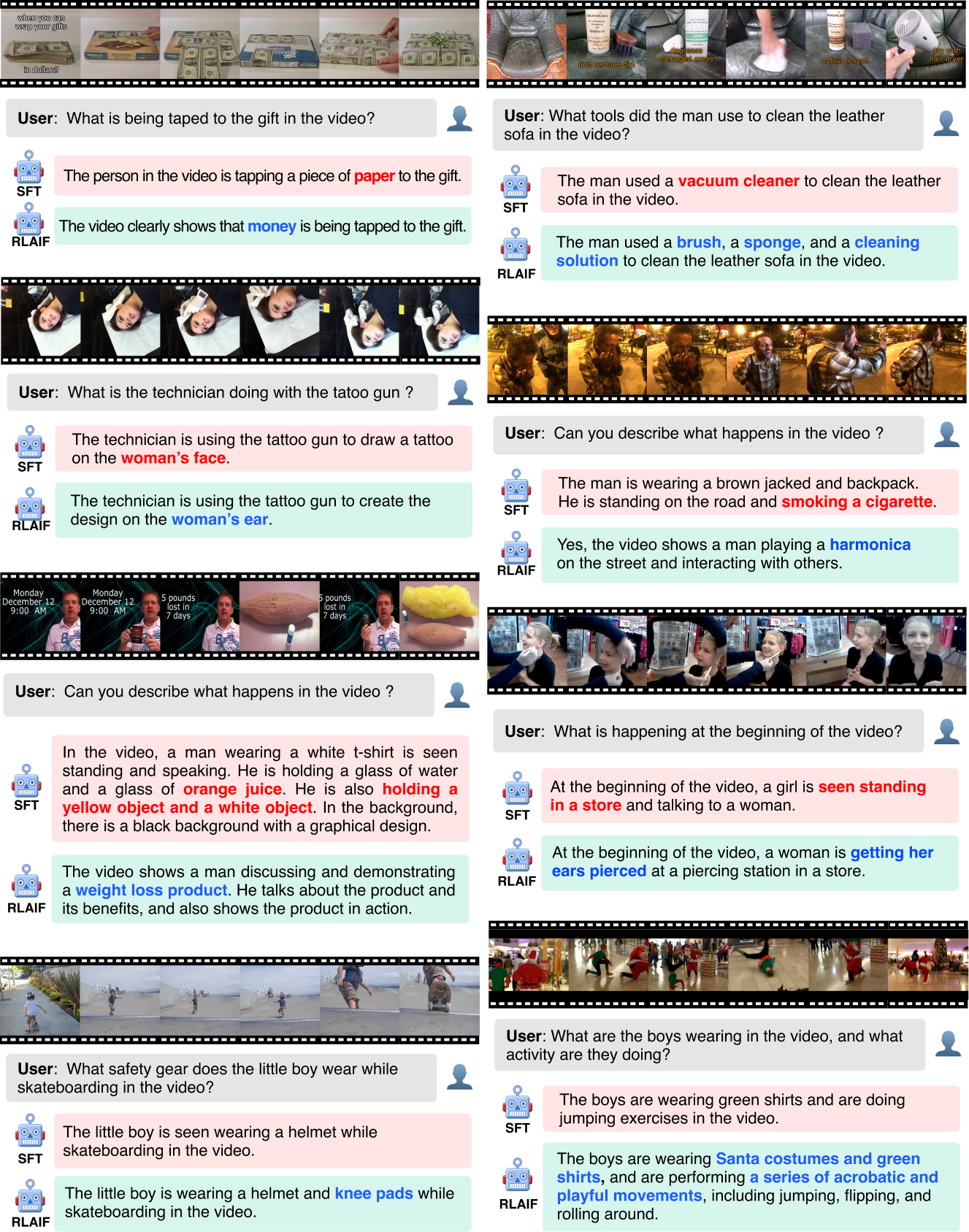}
    \caption{\textbf{More examples of the comparative results between \methodsft and \method.} We present more qualitative examples of questions and answers generated by the \methodsft and \method. The \method yields responses that are more accurately grounded in the visual input, as highlighted in \textcolor{blue}{blue} for accurate responses and \textcolor{red}{red} for less accurate ones from \methodsft.}
    \label{fig:append:more_sft_rlaif}
\end{figure*}

\clearpage

\begin{figure*}[t]
    % \centering
    \includegraphics[width=\linewidth]{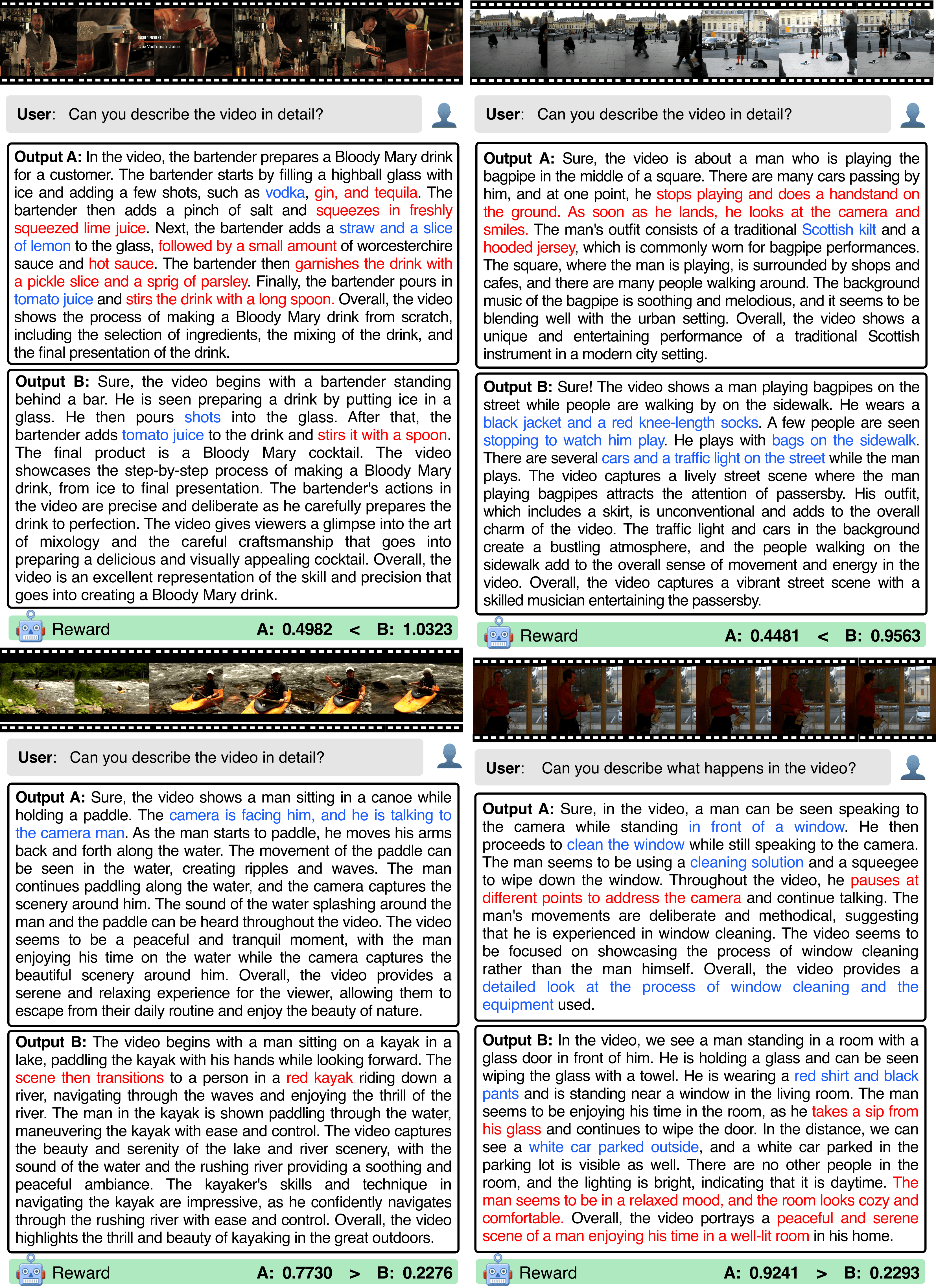}
    \caption{\textbf{Predictions of SFT model and the predicted reward.} We present qualitative examples of questions and two responses generated from the \methodsft, and the reward given to each answer using the trained RM. The reward is higher for the answer with more accurately grounded answers, as highlighted in \textcolor{blue}{blue} for accurate responses and \textcolor{red}{red} for less accurate contents.}
    \label{fig:append:reward_viz}
\end{figure*}

\end{document}